\documentclass[journal]{IEEEtai}

\usepackage[colorlinks,urlcolor=blue,linkcolor=blue,citecolor=blue]{hyperref}

\usepackage{color,array}
\usepackage{stfloats}
\usepackage{bm,bbm,float,booktabs}
\usepackage{multirow}
\usepackage{graphicx}
\usepackage{amsmath,amssymb,amsfonts,amssymb,amsthm}
\usepackage{algorithm,algorithmic}
\newtheorem{theorem}{Theorem}

\newtheorem{lemma}{Lemma}
\newtheorem{definition}{Definition}

\def\Ry{R\'enyi }

\begin{document}
\title{\Ry State Entropy Maximization for Exploration Acceleration in Reinforcement Learning}

\author{Mingqi Yuan, Man-on Pun, \IEEEmembership{Senior Member, IEEE} and Dong Wang
\thanks{This work was supported by National Key Research and Development Program of China under Grant No. 2020YFB1807700. (Corresponding author: Man-On Pun.)}
\thanks{Mingqi Yuan is with the School of Science and Engineering, The Chinese University of Hong Kong, Shenzhen, 518172 China, and also with Shenzhen Research Institute of Big Data, Shenzhen, 518172 China (e-mail: mingqiyuan@link.cuhk.edu.cn).}
\thanks{Man-on Pun is with the School of Science and Engineering, The Chinese University of Hong Kong, Shenzhen, 518172 China, and also with Shenzhen Research Institute of Big Data, Shenzhen, 518172 China (e-mail: simonpun@cuhk.edu.cn).}
\thanks{Dong Wang is with the School of Science and Engineering, The Chinese University of Hong Kong, Shenzhen, 518172 China (e-mail: wangdong@cuhk.edu.cn).}
\thanks{This paragraph will include the Associate Editor who handled your paper.}}

\markboth{Journal of IEEE Transactions on Artificial Intelligence, Vol. 00, No. 0, Month 2020}
{First A. Author \MakeLowercase{\textit{et al.}}: Bare Demo of IEEEtai.cls for IEEE Journals of IEEE Transactions on Artificial Intelligence}

\maketitle

\begin{abstract}
One of the most critical challenges in deep reinforcement learning is to maintain the long-term exploration capability of the agent. To tackle this problem, it has been recently proposed to provide intrinsic rewards for the agent to encourage exploration. However, most existing intrinsic reward-based methods proposed in the literature fail to provide sustainable exploration incentives, a problem known as vanishing rewards. In addition, these conventional methods incur complex models and additional memory in their learning procedures, resulting in high computational complexity and low robustness. In this work, a novel intrinsic reward module based on the \Ry entropy is proposed to provide high-quality intrinsic rewards. It is shown that the proposed method actually generalizes the existing state entropy maximization methods.	In particular, a $k$-nearest neighbor estimator is introduced for entropy estimation while a $k$-value search method is designed to guarantee the estimation accuracy. Extensive simulation results demonstrate that the proposed \Ry entropy-based method can achieve higher performance as compared to existing schemes. The simulation code used in this work is available in the following GitHub website~\footnote{https://github.com/yuanmingqi/RISE}.
\end{abstract}

\begin{IEEEImpStatement}
Reinforcement learning (RL) has demonstrated impressive performance in many complex games such as Go and StarCraft. However, the existing RL algorithms suffer from prohibitively expensive computational complexity, poor generalization ability, and low robustness, which hinders its practical applications in the real world. Thus, it is essential to develop more effective RL algorithms for real-life applications such as autonomous driving and smart manufacturing. To tackle this problem, one critical design challenge is to improve the exploration mechanism of RL to realize efficient policy learning. This work proposes a simple and yet, effective method that can significantly improve the exploration ability of RL algorithms, which can be easily applied to real-life applications. For instance, it will facilitate the development of more powerful autonomous driving systems that can adapt to more complex and challenging environments. Finally, this work is also expected to inspire more subsequent research.
\end{IEEEImpStatement}

\begin{IEEEkeywords}
Reinforcement learning, exploration, intrinsic reward, \Ry entropy.
\end{IEEEkeywords}

\section{Introduction}

\IEEEPARstart{R}{einforcement} learning (RL) algorithms have to be designed to achieve an appropriate balance between exploitation and exploration \cite{sutton2018reinforcement}. However, many existing RL algorithms suffer from insufficient exploration, {\em i.e.} the agent cannot keep exploring the environment to visit all possible state-action pairs \cite{watkins1992q}. As a result, the learned policy prematurely falls into local optima after finite iterations \cite{stadie2015incentivizing}. To address the problem, a simple approach is to employ stochastic policies such as the $\epsilon$-greedy policy and the Boltzmann exploration \cite{lecun2015deep}. These policies randomly select one action with a non-zero probability in each state. For continuous control tasks, an additional noise term can be added to the action to perform limited exploration. Despite the fact that such techniques can eventually learn the optimal policy in the tabular setting, they are futile when handling complex environments with high-dimensional observations.

To cope with the exploration problems above, recent approaches proposed to leverage intrinsic rewards to encourage exploration. In sharp contrast to the extrinsic rewards explicitly given by the environment, intrinsic rewards represent the inherent learning motivation or curiosity of the agent \cite{singh2005intrinsically}. Most existing intrinsic reward modules can be broadly categorized into novelty-based and prediction error-based approaches \cite{xu2017study, csimcsek2006intrinsic, burda2018exploration, lee2019efficient}. For instance, \cite{strehl2008analysis, ostrovski2017count, bellemare2016unifying} employed a state visitation counter to evaluate the novelty of states, and the intrinsic rewards are defined to be inversely proportional to the visiting frequency. As a result, the agent is encouraged to revisit those infrequent states while increasing the probability of exploring new states. In contrast, \cite{pathak2017curiosity, yu2020intrinsic, stadie2015incentivizing} followed an alternative approach in which the prediction error of a dynamic model is utilized as intrinsic rewards. Given a state transition, an auxiliary model was designed to predict a successor state based on the current state-action pair. After that, the intrinsic reward is computed as the Euclidean distance between the predicted and the true successor states. In particular, \cite{burda2018large} attempted to perform RL using only the intrinsic rewards, showing that the agent could achieve considerable performance in many experiments. Despite their good performance, these count-based and prediction error-based methods suffer from vanishing intrinsic rewards, {\em i.e.} the intrinsic rewards decrease with visits \cite{ecoffet2019go}. The agent will have no additional motivation to explore the environment further once the intrinsic rewards decay to zero. To maintain exploration across episodes, \cite{badia2020never} proposed a never-give-up (NGU) framework that learns mixed intrinsic rewards composed of episodic and life-long state novelty. NGU evaluates the episodic state novelty using a slot-based memory and pseudo-count method \cite{bellemare2016unifying}, which encourages the agent to visit as many distinct states as possible in each episode. Since the memory is reset at the beginning of each episode, the intrinsic rewards will not decay during the training process. Meanwhile, NGU further introduced a random network distillation (RND) module to capture the life-long novelty of states, which prevents the agent from visiting familiar states across episodes \cite{burda2018exploration}. However, NGU suffers from complicated architecture and high computational complexity, making it difficult to be applied in arbitrary tasks. A more straightforward framework entitled rewarding-impact-driven-exploration (RIDE) is proposed in \cite{raileanu2020ride}. RIDE inherits the inverse-forward pattern of \cite{pathak2017curiosity}, in which two dynamic models are leveraged to reconstruct the transition process. More specifically, the Euclidean distance between two consecutive encoded states is utilized as the intrinsic reward, which encourages the agent to take actions that result in more state changes. Moreover, RIDE uses episodic state visitation counts to discount the generated rewards, preventing the agent from staying at states that lead to large embedding differences while avoiding the television dilemma reported in \cite{savinov2018episodic}. 

However, both NGU and RIDE pay excessive attention to specific states while failing to reflect the global exploration extent. Furthermore, they suffer from poor mathematical interpretability and performance loss incurred by auxiliary models. To circumvent these problems, \cite{seo2021state} proposed a state entropy maximization method entitled random-encoder-for-efficient-exploration (RE3), forcing the agent to visit the state space more equitably. In each episode, the observation data is collected and encoded using a randomly initialized deep neural network. After that, a $k$-nearest neighbor estimator is leveraged to realize efficient entropy estimation \cite{singh2003nearest}. Simulation results demonstrated that RE3 significantly improved the sampling efficiency of both model-free and model-based RL algorithms at the cost of less computational complexity. Despite its many advantages, RE3 ignores the important $k$-value selection while its default random encoder entails low adaptability and robustness. Furthermore, \cite{zhang2021exploration} found that the Shannon entropy-based objective function may lead to a policy that visits some states with a vanishing probability, and proposed to maximize the \Ry entropy of state-action distribution (Max\Ry). In contrast to RE3, RISE provides a more appropriate optimization objective for sustainable exploration. However, \cite{zhang2021exploration} leverages a variation auto-encoder (VAE) to estimate the state-action distribution, which produces high computational complexity and may mislead the agent due to imperfect estimation \cite{kingma2013auto}.

Inspired by the discussions above, we propose to devise a more efficient and robust method for state entropy maximization to improve exploration in RL. In this paper, we propose a \textbf{R}\'eny\textbf{I} \textbf{S}tate \textbf{E}ntropy (RISE) maximization framework to provide high-quality intrinsic rewards. Our main contributions are summarized as follows:

\begin{itemize}
	\item We propose a \Ry entropy-based intrinsic reward module that generalizes the existing state entropy maximization methods such as RE3, and provide theoretical analysis for the \Ry entropy-based learning objective. The new module can be applied in arbitrary tasks with significantly improved exploration efficiency for both model-based and model-free RL algorithms;
	
	\item By leveraging (VAE) model, the proposed module can realize efficient and robust encoding operation for accurate entropy estimation, which guarantees its generalization capability and adaptability. Moreover, a search algorithm is devised for the $k$-value selection to reduce the uncertainty of performance loss caused by random selection;
	
	\item Finally, extensive simulation is performed to compare the performance of RISE against existing methods using both discrete and continuous control tasks as well as several hard exploration games. Simulation results confirm that the proposed module achieve superior performance with higher efficiency.
\end{itemize}

\section{Problem Formulation}\label{section:pf}
We study the following RL problem that considers a Markov decision process (MDP) characterized by a tuple $\mathcal{M}=\langle \mathcal{S},\mathcal{A},\mathcal{T},r,\rho({\bm s}_0),\gamma\rangle$ \cite{sutton2018reinforcement}, in which $\mathcal{S}$ is the state space, $\mathcal{A}$ is the action space, $\mathcal{T}({\bm s}'|{\bm s},{\bm a})$ is the transition probability, $r({\bm s},{\bm a}):\mathcal{S}\times\mathcal{A}\rightarrow\mathbb{R}$ is the reward function, $\rho({\bm s}_0)$ is the initial state distribution, and $\gamma\in(0,1]$ is a discount factor, respectively. We denote by $\pi({\bm a}|{\bm s})$ the policy of the agent that observes the state of the environment before choosing an action from the action space. The objective of RL is to find the optimal policy $\pi^{*}$ that maximizes the expected discounted return given by
\begin{equation}\label{eq:rl objective}
\pi^{*}=\underset{\pi\in\Pi}{\rm argmax}\;\mathbb{E}_{\tau\sim\pi}\sum_{t=0}^{T-1}\gamma^{t}r_{t}({\bm s}_t,{\bm a}_t),
\end{equation}
where $\Pi$ is the set of all stationary policies, and $\tau=({\bm s}_{0},{\bm a}_{0},\dots,{\bm a}_{T-1},{\bm s}_{T})$ is the trajectory collected by the agent.

In this paper, we aim to improve the exploration in RL. To guarantee the completeness of exploration, the agent is required to visit all possible states during training. Such an objective can be regarded as the Coupon collector's problem conditioned upon a nonuniform probability distribution \cite{flajolet1992birthday}, in which the agent is the collector and the states are the coupons. Denote by $d^{\pi}(\bm{s})$ the state distribution induced by the policy $\pi$. Assuming that the agent takes $\tilde{T}$ environment steps to finish the collection, we can compute the expectation of $\tilde{T}$ as
\begin{equation}\label{eq:Epi}
\mathbb{E}_{\pi}(\tilde{T})=\int_{0}^{\infty}\bigg( 1-\prod_{i=1}^{|\mathcal{S}|}(1-e^{-d^{\pi}(\bm{s}_{i})t}) \bigg)\,dt,
\end{equation}
where $\left|\cdot\right|$ stands for the cardinality of the enclosed set $\mathcal{S}$.

For simplicity of notation, we sometimes omit the superscript in $d^{\pi}(\bm{s})$ in the sequel. Efficient exploration aims to find a policy that optimizes $\min_{\pi\in\Pi}\:\mathbb{E}_{\pi}(\tilde{T})$. However, it is non-trivial to evaluate Eq.~\eqref{eq:Epi} due to the improper integral, not to mention solving the optimization problem. To address the problem, it is common to leverage the Shannon entropy to make a tractable objective function, which is defined as
\begin{equation}\label{eq:Hd}
H(d)=-\mathbb{E}_{\bm{s}\sim d(\bm{s})}\big[\log d({\bm s})\big].
\end{equation}
However, this objective function may lead to a policy that visits some states with a vanishing probability. In the following section, we will first employ a representative example to demonstrate the practical drawbacks of Eq.~\eqref{eq:Hd} before introducing the \Ry entropy to address the problem.

\begin{figure*}[t]
	\centering
	\includegraphics[width=0.90\linewidth]{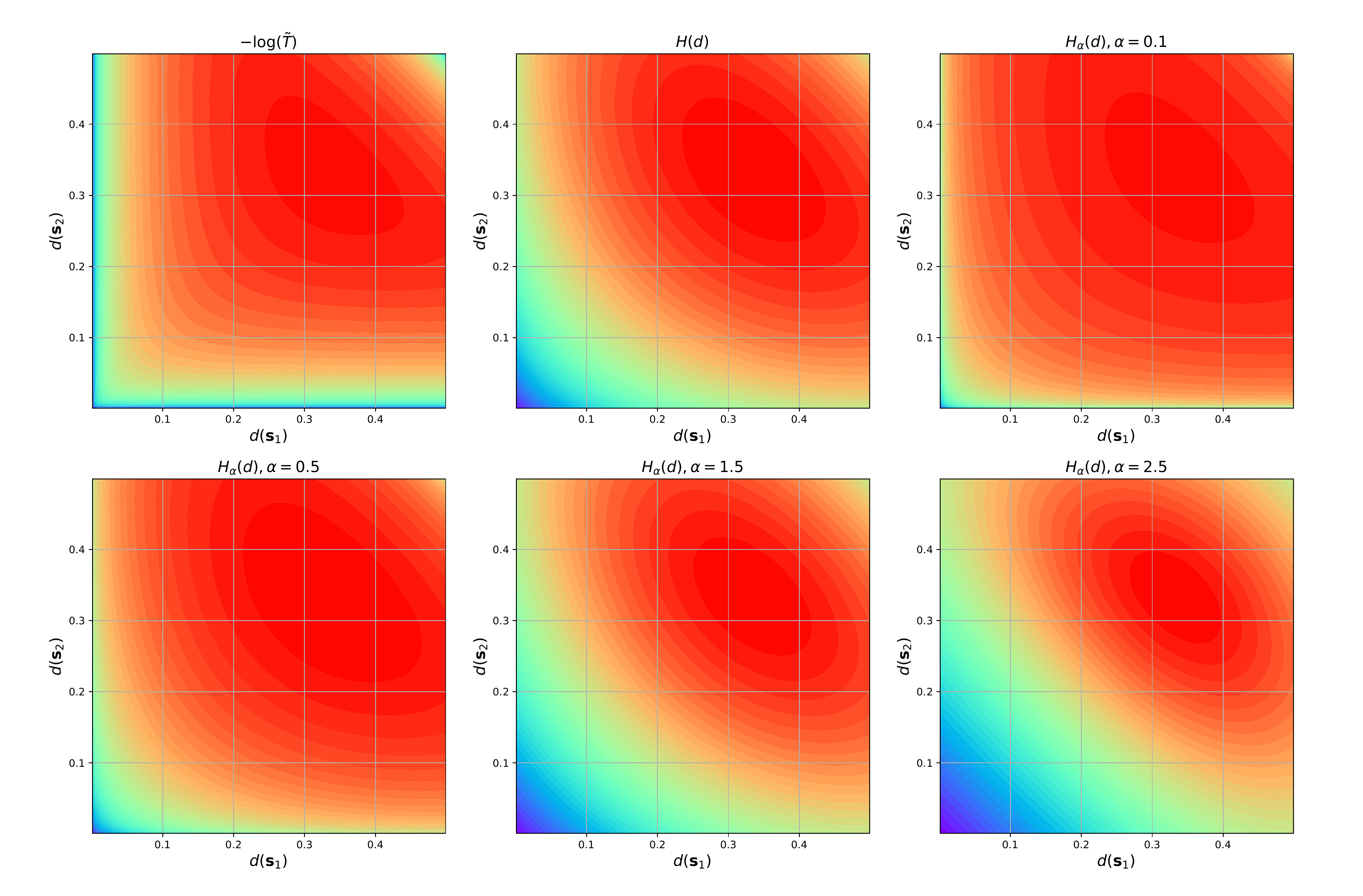}
	\caption{The contours of different objective functions when $|\mathcal{S}|=3$.}
	\label{fig:contour}
\end{figure*}

\section{\Ry State Entropy Maximization}
\subsection{\Ry State Entropy}
We first formally define the \Ry entropy as follows:
\begin{definition}[\Ry Entropy]\label{def:re}
	Let $X\in\mathbb{R}^{m}$ be a random vector that has a density function $f(\bm{x})$ with respect to Lebesgue measure on $\mathbb{R}^{m}$, and let $\mathcal{X}=\{\bm{x}\in\mathbb{R}^{m}:f(\bm{x})>0\}$ be the support of the distribution. The \Ry entropy of order $\alpha\in(0,1)\cup(1,+\infty)$ is defined as \cite{zhang2021exploration}:
	\begin{equation}\label{eq:hf}
	H_{\alpha}(f)=\frac{1}{1-\alpha}\log \int_{\mathcal{X}}f^{\alpha}(\bm{x})d \bm{x}.
	\end{equation}
\end{definition}

Using Definition~\ref{def:re}, we propose the following \Ry state entropy (RISE):
\begin{equation}
	H_{\alpha}(d)=\frac{1}{1-\alpha}\log \int_{\mathcal{S}}d^{\alpha}(\bm{s})d\bm{s}.
\end{equation}

Fig.~\ref{fig:contour} use a toy example to visualize the contours of different objective functions when an agent learns from an environment characterized by only three states. As shown in Fig.~\ref{fig:contour}, $-\log(\tilde{T})$ decreases rapidly when any state probability approaches zero, which prevents the agent from visiting a state with a vanishing probability while encouraging the agent to explore the infrequently-seen states. In contrast, the Shannon entropy remains relatively large as the state probability approaches zero. Interestingly, Fig.~\ref{fig:contour} shows that this problem can be alleviated by the \Ry entropy as it better matches $-\log(\tilde{T})$. The Shannon entropy is far less aggressive in penalizing small probabilities, while the \Ry entropy provides more flexible exploration intensity. 

\subsection{Theoretical Analysis}
To maximize $H_{\alpha}(d)$, we consider using a maximum entropy policy computation (MEPC) algorithm proposed by \cite{hazan2019provably}, which uses the following two oracles: 
\begin{definition}[Approximating planning oracle]
	Given a reward function $r:\mathcal{S}\rightarrow\mathbb{R}$ and a gap $\epsilon$, the planning oracle returns a policy by $\pi=O_{\rm AP}(r,\epsilon_1)$, such that
	\begin{equation}
		V^{\pi}\geq\max_{\pi\in\Pi}V^{\pi}-\epsilon,
	\end{equation}
where $V^{\pi}$ is the state-value function.
\end{definition}
\begin{definition}[State distribution estimation oracle]
	Given a gap $\epsilon$ and a policy $\pi$, this oracle estimates the state distribution by $\hat{d}=O_{\rm DE}(\pi,\epsilon)$, such that
	\begin{equation}
		\Vert d-\hat{d}\Vert_{\infty}\leq\epsilon.
	\end{equation}
\end{definition}
Given a set of stationary policies $\hat{\Pi}=\{\pi_0,\pi_1,\dots\}$, we define a mixed policy as $\pi_{\rm mix}=(\bm{\omega},\hat{\Pi})$, where $\omega$ contains the weighting coefficients. Then the induced state distribution is
\begin{equation}
	d^{\pi_{\rm mix}}=\sum_{i}\bm{\omega}_{i}d^{\pi_i}(\bm{s}).
\end{equation}
Finally, the workflow of MEPC is summarized in Algorithm \ref{algo:mpec}.

\begin{algorithm}[h]
	\caption{MEPC}
	\label{algo:mpec}
	\begin{algorithmic}[1]
		\STATE Set number of iterations $T$, step size $\eta$, planning oracle error tolerance $\epsilon_1>0$, and state distribution oracle error error tolerance $\epsilon_2>0$;
		\STATE Set $\hat{\Pi}_0=\{\pi_0\}$, where $\pi_0$ is an arbitrary policy;
		\STATE Set $\omega_0=1$;
		\FOR {$t=0,\dots,T-1$}
		\STATE Invoke the state distribution oracle on $\pi_{{\rm mix}, t}=(\omega,\hat{H}_t)$:
		\begin{equation}\nonumber
			\hat{d}_{{\rm mix}, t}=O_{\rm AP}(r,\epsilon_1);
		\end{equation}

		\STATE Define the reward function $r_{t}$ as
		\begin{equation}\nonumber
			r_{t}(\bm{s})=\nabla H_{\alpha}(\hat{d}_{{\rm mix}, t});
		\end{equation}
	
		\STATE Approximate the optimal policy on $r_t$:
		\begin{equation}\nonumber
			\pi_{t+1}=O_{\rm DE}(\pi,\epsilon_2);
		\end{equation}
	
		\STATE Update $\pi_{{\rm mix}, t}=(\omega_{t+1},\hat{\Pi})$ by:
		\begin{equation}\nonumber
			\begin{aligned}
				\hat{\Pi}_{t+1}=(\pi_0,\dots,\pi_t,\pi_{t+1})\\
				\omega_{t+1}=((1-\eta)\omega_{t},\eta);
			\end{aligned}
		\end{equation}
	
		\ENDFOR
		\STATE Return $\pi_{{\rm mix}, T}=(\omega_T,\hat{\Pi}_T)$.
	\end{algorithmic}
\end{algorithm}

Consider the discrete case of \Ry state entropy and set $\alpha\in(0,1)$, we have 
\begin{equation}
	H_{\alpha}(d)=\frac{1}{1-\alpha}\log\sum_{\bm{s}\in\mathcal{S}}d^{\alpha}(\bm {s}).
\end{equation}
To maximize $H_{\alpha}(d)$, we can alternatively maximize
\begin{equation}
	\tilde{H}_{\alpha}(d)=\frac{1}{1-\alpha}\sum_{\bm{s}\in\mathcal{S}}d^{\alpha}(\bm {s}).
\end{equation}
Since $\tilde{H}_{\alpha}(d)$ is not smooth, we may consider a smoothed $\tilde{H}_{\alpha,\sigma}(d)$ defined as
\begin{equation}
	\tilde{H}_{\alpha,\sigma}(d)=\frac{1}{1-\alpha}\sum_{\bm{s}\in\mathcal{S}}(d(\bm {s})+\sigma)^{\alpha},
\end{equation}
where $\sigma>0$.
\begin{lemma}\label{lemma:smooth}
	$\tilde{H}_{\alpha,\sigma}(d)$ is $\beta$-smooth, such that
	\begin{equation}
		\Vert\nabla\tilde{H}_{\alpha,\sigma}(d)-\nabla\tilde{H}_{\alpha,\sigma}(d')\Vert_{\infty}\leq \beta\Vert d-d'\Vert_{\infty},
	\end{equation}
where $\beta=\alpha\sigma^{\alpha-2}$.
\end{lemma}
\begin{proof}
	See proof in Appendix \ref{appendix:proof of lemma}.
\end{proof}

Now we are ready to give the following theorem:
\begin{theorem}\label{theorem:sample complexity}
	For any $\epsilon>0$, let $\epsilon_1=0.1\epsilon, \epsilon_2=0.1\beta^{-1}\epsilon$ and $\eta=0.1\beta^{-1}\epsilon$. It holds
	\begin{equation}
		\tilde{H}_{\alpha,\sigma}(d^{\pi_{{\rm mix},T}})\geq \max_{\pi\in\Pi}\tilde{H}_{\alpha,\sigma}(d^{\pi})-\epsilon,
	\end{equation}
	if Algorithm \ref{algo:mpec} is run for 
	\begin{equation}
		T\geq \frac{10\alpha\sigma^{\alpha-2}}{\epsilon}\log\frac{10\alpha\sigma^{\alpha-1}}{(1-\alpha)\epsilon}.
	\end{equation}
\end{theorem}
\begin{proof}
	See proof in Appendix \ref{appendix:proof of theorem}.
\end{proof}
Theorem \ref{theorem:sample complexity} demonstrates the computational complexity when using MEPC to maximize $\tilde{H}$. Moreover, a small $\alpha$ will contribute to the exploration phase, which is consistent with the analysis in \cite{zhang2021exploration}.

\subsection{Fast Entropy Estimation}

However, it is non-trivial to apply MEPC when handling complex environments with high-dimensional observations. To address the problem, we propose to utilize the following $k$-nearest neighbor estimator to realize efficient estimation of the \Ry entropy \cite{leonenko2008class}. Note that $\pi$ in Eq.~\eqref{eq:estimator} denotes the ratio between the circumference of a circle to its diameter.

\begin{theorem}[Estimator]\label{def:re estimation}
	Denote by  $\{X_{i}\}_{i=1}^{N}$ a set of independent random vectors from the distribution $X$. For $k<N,k\in\mathbb{N}$, $\tilde{X}_{i}$ stands for the $k$-nearest neighbor of $X_{i}$ among the set. We estimate the \Ry entropy using the sample mean as follows:
	\begin{equation}\label{eq:estimator}
	\hat{H}^{k,\alpha}_{N}(f)=\frac{1}{N}\sum_{i=1}^{N}\big[ (N-1)V_{m}C_{k}\Vert X_{i}-\tilde{X}_{i} \Vert^{m} \big]^{1-\alpha},
	\end{equation}
	where $C_{k}=\bigg[ \frac{\Gamma(k)}{\Gamma(k+1-\alpha)} \bigg]^{\frac{1}{1-\alpha}}$ and $V_{m}=\frac{\pi^{\frac{m}{2}}}{\Gamma(\frac{m}{2}+1)}$ is the volume of the unit ball in $\mathbb{R}^{m}$, and $\Gamma(\cdot)$ is the Gamma function, respectively. Moreover, it holds
	\begin{equation}
		\lim_{N\rightarrow\infty}\hat{H}^{k,\alpha}_{N}(f)=H_{\alpha}(f).
	\end{equation}
\end{theorem}
\begin{proof}
	See proof in \cite{leonenko2008class}.
\end{proof}

\begin{figure*}[ht]
	\centering
	\includegraphics[width=0.8\linewidth]{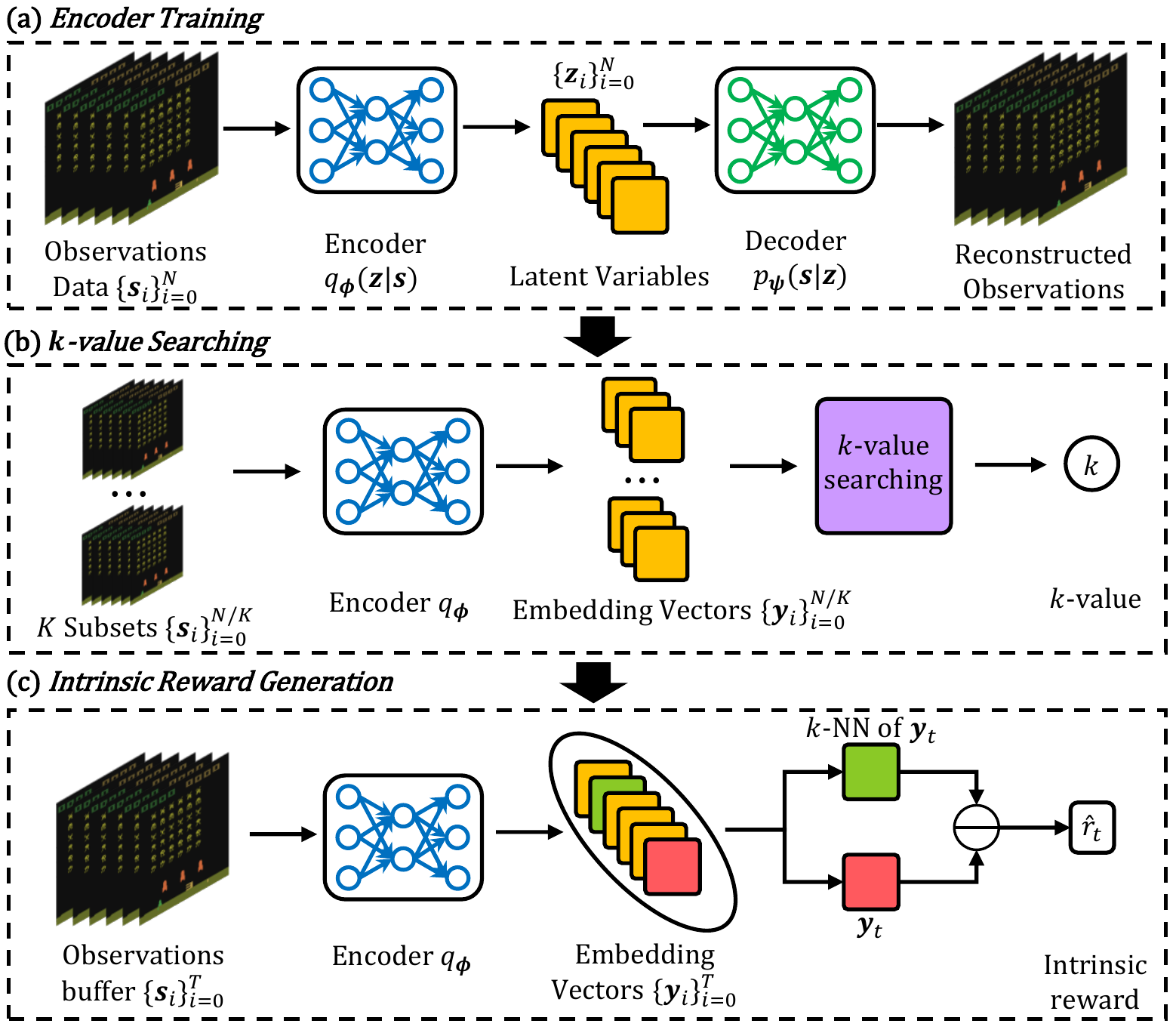}
	\caption{The overview architecture of RISE. (a) VAE model for embedding observations. (b) $k$-value searching. (c) The generation of intrinsic rewards, where $k$-NN is $k$-nearest neighbor and $\ominus$ denotes the Euclidean distance.}
	\label{fig:rise}
\end{figure*}

Given a trajectory $\tau=\{\bm{s}_{0},\bm{a}_{0},\dots,\bm{a}_{T-1},\bm{s}_{T}\}$ collected by the agent, we approximate the \Ry state entropy in Eq.~\eqref{eq:hf} using Eq.~\eqref{eq:estimator} as
\begin{equation}
\begin{aligned}
\hat{H}^{k,\alpha}_{T}(d)&=\frac{1}{T}\sum_{i=0}^{T-1}\big[ (T-1) V_{m}C_{k}\Vert \bm{y}_{i}-\tilde{\bm{y}}_{i} \Vert^{m} \big]^{1-\alpha},\\
&\propto\frac{1}{T}\sum_{i=0}^{T-1} \Vert \bm{y}_{i}-\tilde{\bm{y}}_{i} \Vert^{1-\alpha},
\end{aligned}
\end{equation}
where $\bm{y}_{i}$ is the encoding vector of $\bm{s}_i$ and $\tilde{\bm{y}}_{i}$ is the $k$-nearest neighbor of $\bm{y}_{i}$. After that, we define the intrinsic reward that takes each transition as a particle:
\begin{equation}\label{eq:intrinsic reward}
\hat{r}(\bm{s}_{i})=\Vert \bm{y}_{i}-\tilde{\bm{y}}_{i} \Vert^{1-\alpha},
\end{equation}
where $\hat{r}(\cdot)$ is used to distinguish the intrinsic reward from the extrinsic reward $r(\cdot)$. Eq.~\eqref{eq:intrinsic reward} indicates that the agent needs to visit as more distinct states as possible to obtain higher intrinsic rewards.

Such an estimation method requires no additional auxiliary models, which significantly promotes the learning efficiency. Equipped with the intrinsic reward, the total reward of each transition $(\bm{s}_t,\bm{a}_t,\bm{s}_{t+1})$ is computed as
\begin{equation}
	r^{\rm total}_t=r({\bm s}_t,{\bm a}_t)+\lambda_{t}\cdot\hat{r}(\bm{s}_t)+\zeta\cdot H(\pi(\cdot|\bm{s}_t)),
\end{equation}
where $H(\pi(\cdot|\bm{s}_t))$ is the action entropy regularizer for improving the exploration on action space, $\lambda_{t}=\lambda_{0}(1-\kappa)^{t}$ and $\zeta$ are two non-negative weight coefficients, and $\kappa$ is a decay rate. 


\section{Robust Representation Learning}
While the \Ry state entropy encourages exploration in high-dimensional observation spaces, several implementation issues have to be addressed in its practical deployment. First of all, observations have to be encoded into low-dimensional vectors in calculating the intrinsic reward. While a randomly initialized neural network can be utilized as the encoder as proposed in \cite{seo2021state}, it cannot handle more complex and dynamic tasks, which inevitably incurs performance loss. Moreover, since it is less computationally expensive to train an encoder than RL, we propose to leverage the VAE to realize efficient and robust embedding operation, which is a powerful generative model based on the Bayesian inference \cite{kingma2013auto}. As shown in Fig.~\ref{fig:rise}(a), a standard VAE is composed of a recognition model and a generative model. These two models represent a probabilistic \textit{encoder} and a probabilistic \textit{decoder}, respectively.

We denote by $q_{\bm \phi}(\bm{z}|\bm{s})$ the recognition model represented by a neural network with parameters $\bm{\phi}$. The recognition model accepts an observation input before encoding the input into latent variables. Similarly, we represent the generative model as $p_{\bm \psi}(\bm{s}|\bm{z})$ using a neural network with parameters $\bm{\psi}$, accepting the latent variables and reconstructing the observation. Given a trajectory $\tau=\{\bm{s}_{0},\bm{a}_{0},\dots,\bm{a}_{T-1},\bm{s}_{T}\}$, the VAE model is trained by minimizing the following loss function:
\begin{equation}\label{eq:vae loss}
\begin{aligned}
L(\bm{s}_{t};{\bm \phi},{\bm \psi})&=\mathbb{E}_{q_{\bm \phi}(\bm{z}|\bm{s}_{t})}\big[\log p_{\bm \psi}(\bm{s}_{t}|\bm{z})\big]\\
&-D_{{\rm KL}}\big(q_{\bm\phi}(\bm{z}|\bm{s}_{t})\Vert p_{\bm \psi}(\bm{z})\big),
\end{aligned}
\end{equation}
where $t=0,\dots,T$, $D_{{\rm KL}}(\cdot)$ is the Kullback-Liebler (KL) divergence.

Next, we will elaborate on the design of the $k$ value to improve the estimation accuracy of the state entropy. \cite{singh2003nearest} investigated the performance of this entropy estimator for some specific probability distribution functions such as uniform distribution and Gaussian distribution. Their simulation results demonstrated that the estimation accuracy first increased before decreasing as the $k$ value increases. To circumvent this problem, we propose our $k$-value searching scheme as shown in Fig.~\ref{fig:rise}(b). We first divide the observation dataset into $K$ subsets before the encoder encodes the data into low-dimensional embedding vectors. Assuming that all the data samples are independent and identically distributed, an appropriate $k$ value should produce comparable results on different subsets. By exploiting this intuition, we propose to search the optimal $k$ value that minimizes the min-max ratio of entropy estimation set. Denote by $\pi_{\bm{\theta}}$ the policy network, the detailed searching algorithm is summarized in Algorithm~\ref{algo:k search}.
\begin{algorithm}[h]
	\caption{$k$-value searching method}
	\label{algo:k search}
	\begin{algorithmic}[1]
		\STATE Initialize a policy network $\pi_{\bm{\theta}}$;
		\STATE Initialize the number of sample steps $N$, the threshold $k_{\rm max}$ of $k$, a null array $\delta$ with length $k_{\rm max}$, and the number of subsets $K$;
		\STATE Execute policy $\pi_{\bm{\theta}}$ and collect the trajectory $\tau=\{\bm{s}_{0},\bm{a}_{0},\dots,\bm{a}_{N-1},\bm{s}_{N}\}$;
		\STATE Divide the observations dataset $\{\bm{s}_{i}\}_{i=0}^{N}$ into $K$ subsets randomly;
		\FOR {$k=1,2,\dots,k_{\rm max}$}
		\STATE Calculate the estimated entropy on $K$ subsets using Eq.~\eqref{eq:estimator}:
		\begin{equation}
		\hat{\bm{H}}_{k}=(\hat{H}_{k, N/K, \alpha}[1],\dots,\hat{H}_{k, N/K, \alpha}[K])
		\end{equation}
		
		\STATE Calculate the min-max ratio $\delta(\hat{\bm{H}}_{k})$ and $\bm{\delta}[k] \leftarrow \delta(\hat{\bm{H}}_{k})$;
		\ENDFOR
		\STATE Output $k=\underset{k}{\rm argmin}\:\bm{\delta}[k]$.
	\end{algorithmic}
\end{algorithm}

Finally, we are ready to propose our RISE framework by exploiting the optimal $k$ value derived above. As shown in Fig.~\ref{fig:rise}(c), the proposed RISE framework first encodes the high-dimensional observation data into low-dimensional embedding vectors through $q:\mathcal{S}\rightarrow\mathbb{R}^{m}$. After that, the Euclidean distance between $\bm{y}_{t}$ and its $k$-nearest neighbor is computed as the intrinsic reward. Algorithm~\ref{algo:rise on-policy} and Algorithm~\ref{algo:rise off-policy} summarize the on-policy and off-policy RL versions of the proposed RISE, respectively. In the off-policy version, the entropy estimation is performed on the sampled transitions in each step. As a result, a larger batch size can improve the estimation accuracy. It is worth pointing out that RISE can be straightforwardly integrated into any existing RL algorithms such as Q-learning and soft actor-critic, providing high-quality intrinsic rewards for improved exploration.

\begin{algorithm}[h]
	\caption{On-policy RL Version of RISE}
	\label{algo:rise on-policy}
	\begin{algorithmic}[1]
		\STATE \textbf{Phase 1: $k$-value searching and encoder training}
		\STATE Initialize the policy network $\pi_{\bm{\theta}}$, encoder $q_{\bm \phi}$ and decoder $p_{\bm \psi}$;
		\STATE Initialize the number of sample steps $N$, the threshold $k_{\rm max}$ of $k$-value, the embedding size $m$ and the number subsets $K$ ;
		\STATE Execute policy and collect observations data $\{\bm{s}_{i}\}_{i=1}^{N}$;
		\STATE Use $\{\bm{s}_{i}\}_{i=1}^{N}$ to train the encoder;
		\STATE Use Algorithm \ref{algo:k search} to select $k$-value;
		\STATE \textbf{Phase 2: Policy update}
		\STATE Initialize the maximum episodes $E$, order $\alpha$, coefficients $\lambda_{0},\zeta$ and decay rate $\kappa$;
		\FOR {episode $\ell=1,\dots,E$}
		\STATE Collect the trajectory $\tau_{\ell}=\{\bm{s}_{0},\bm{a}_{0},\dots,\bm{a}_{T-1},\bm{s}_{T}\}$;
		\STATE Compute the embedding vectors $\{\bm{z}_{t}\}_{t=0}^{T}$ of $\{\bm{s}_{t}\}_{t=0}^{T}$ using the encoder $q_{\bm \phi}$;
		\STATE Compute $\hat{r}(\bm{s}_{t})\leftarrow (\Vert \bm{z}_{t}-\tilde{\bm{z}}_{t} \Vert_{2})^{1-\alpha}$;
		\STATE Update $\lambda_{\ell}=\lambda_{0}(1-\kappa)^{\ell}$;
		\STATE Let $r^{\rm total}_{t}=r(\bm{s}_{t},\bm{a}_{t})+\lambda_{\ell} \cdot \hat{r}(\bm{s}_{t})+\zeta\cdot H(\pi(\cdot|\bm{s}_t))$;
		\STATE Update the policy network with transitions $\{\bm{s}_{t},\bm{a}_{t},\bm{s}_{t+1},r^{\rm total}_{t}\}_{t=0}^{T}$ using any on-policy RL algorithms.
		\ENDFOR
	\end{algorithmic}
\end{algorithm}

\section{Experiments}
In this section, we will evaluate our RISE framework on both the tabular setting and environments with high-dimensional observations. We compare RISE against two representative intrinsic reward-based methods, namely RE3 and Max\Ry. A brief introduction of these benchmarking methods can be found in Appendix \ref{appendix:benchmarks}. We also train the agent without intrinsic rewards for ablation studies. As for hyper-parameters setting, we only report the values of the best experiment results.

\subsection{Maze Games}
\begin{figure}[h]
	\centering
	\includegraphics[width=0.5\linewidth]{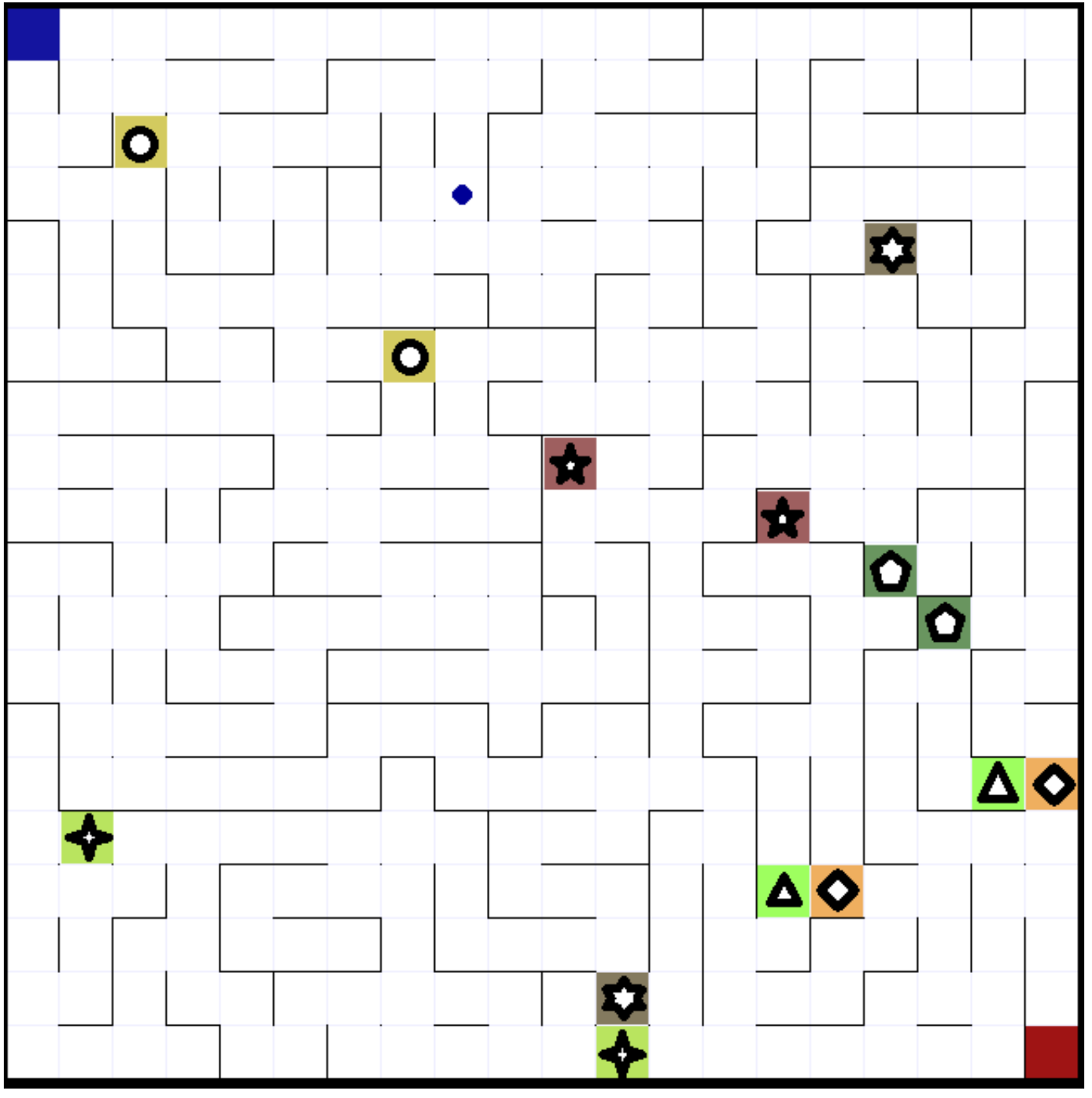}
	\caption{A maze game with grid size $20\times20$.}
	\label{fig:maze}
\end{figure}

In this section, we first leverage a simple but representative example to highlight the effectiveness of the \Ry state entropy-driven exploration. We introduce a grid-based environment Maze2D \cite{matthew2016github} illustrated in Fig.~\ref{fig:maze}. The agent can move one position at a time in one of the four directions, namely left, right, up, and down. The goal of the agent is to find the shortest path from the start point to the end point. In particular, the agent can teleport from a portal to another identical mark.

\subsubsection{Experimental Setting}
The standard Q-learning (QL) algorithm \cite{watkins1992q} is selected as the benchmarking method. We perform extensive experiments on three mazes with different sizes. Note that the problem complexity increases exponentially with the maze size. In each episode, the maximum environment step size was set to $10M^{2}$, where $M$ is the maze size. We initialized the Q-table with zeros and updated the Q-table in every step for efficient training. The update formulation is given by:
\begin{equation}
Q(\bm{s},\bm{a})\leftarrow Q(\bm{s},\bm{a})+\eta[r+\gamma \max_{\bm{a}'}Q(\bm{s}',\bm{a}')-Q(\bm{s},\bm{a})],
\end{equation}
where $Q(\bm{s},\bm{a})$ is the action-value function. The step size was set to $0.2$ while a $\epsilon$-greedy policy with an exploration rate of $0.001$ was employed.

\subsubsection{Performance Comparison}
\begin{figure}[h]
	\centering
	\includegraphics[width=0.95\linewidth]{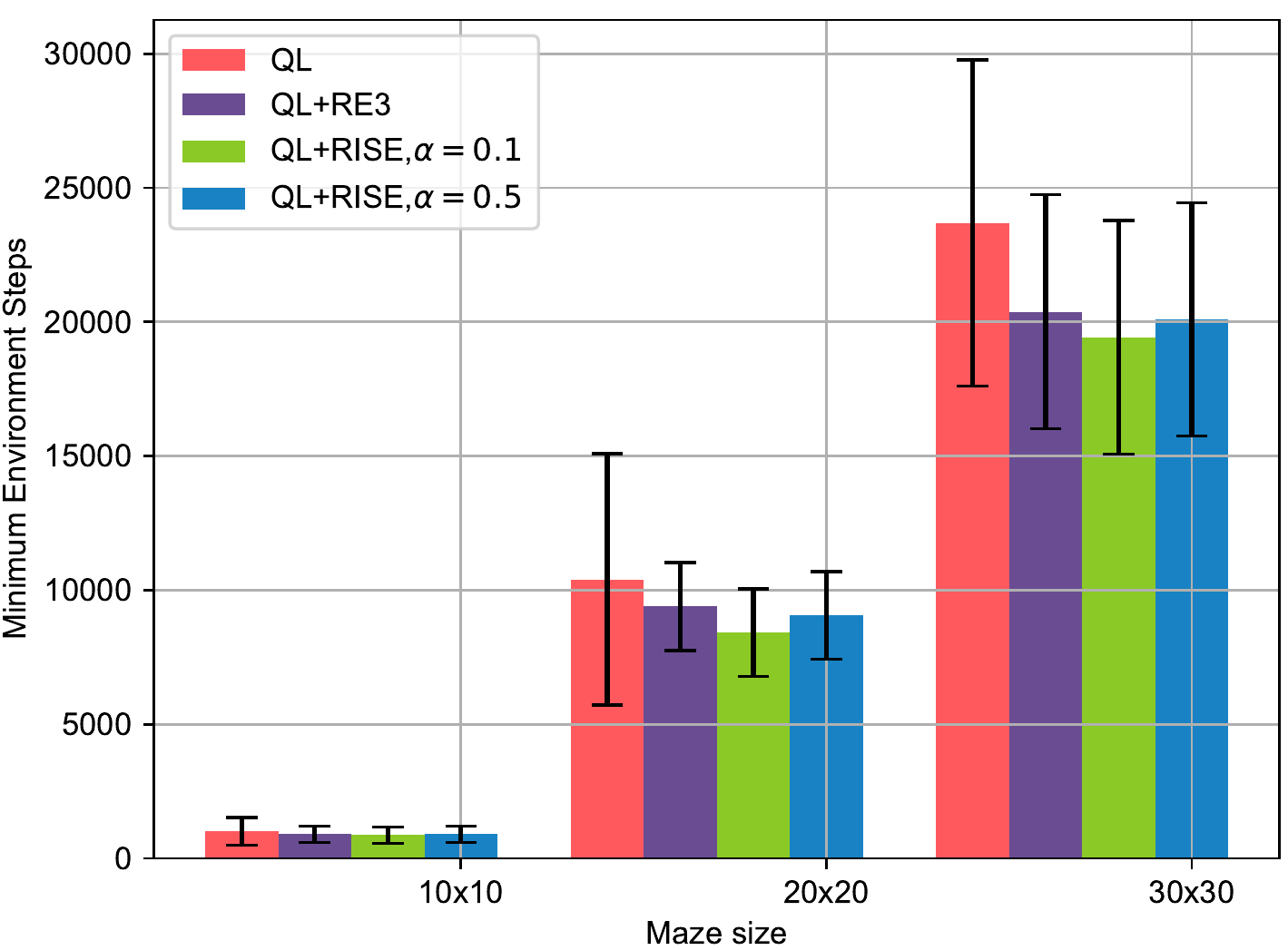}
	\caption{Average exploration performance comparison over $100$ simulation runs.}
	\label{fig:maze return}
\end{figure}

To compare the exploration performance, we choose the minimum number of environment steps taken to visit all states as the key performance indicator (KPI). For instance, a $10\times10$ maze of $100$ grids corresponds to $100$ states. The minimum number of steps for the agent to visit all the possible states is evaluated as its exploration performance. As seen in Fig.~\ref{fig:maze return}, the proposed \textit{Q-learning+RISE} achieved the best performance in all three maze games. Moreover, RISE with smaller $\alpha$ takes less steps to finish the exploration phase. This experiment confirmed the great capability of the \Ry state entropy-driven exploration.

\subsection{Atari Games}

Next, we will test RISE on the Atari games with a discrete action space, in which the player aims to achieve more points while remaining alive \cite{brockman2016openai}. To generate the observation of the agent, we stacked four consecutive frames as one input. These frames were cropped to the size of $(84, 84)$ to reduce the required computational complexity. 


\subsubsection{Experimental Setting}
\begin{table}[h]
	\caption{The CNN-based network architectures.}
	\label{tb:cnn na}
	\centering
	\begin{tabular}{l|l|l}
		\hline
		\textbf{Module} & Policy network $\pi_{\bm \theta}$                                                                                                                                                                                                 & Value network                                                                                                                                                                    \\ \hline
		Input  & States                                                                                                                                                                                                & States                                                                                                                                                                           \\ \hline
		Arch.  & \begin{tabular}[c]{@{}l@{}}8$\times$8 Conv 32, ReLU\\ 4$\times$4 Conv 64, ReLU\\ 3$\times$3 Conv 32, ReLU\\ Flatten\\ Dense 512, ReLU\\ Dense $|\mathcal{A}|$\\ Categorical Distribution\end{tabular} & \begin{tabular}[c]{@{}l@{}}8$\times$8 Conv 32, ReLU\\ 4$\times$4 Conv 64, ReLU\\ 3$\times$3 Conv 32, ReLU\\ Flatten\\ Dense 512, ReLU\\ Dense 1\end{tabular}                     \\ \hline
		Output & Actions                                                                                                                                                                                               & Predicted values                                                                                                                                                                           \\ \hline
		\textbf{Module} & Encoder $p_{\bm \psi}$                                                                                                                                                                                                       & Decoder $q_{\bm \phi}$                                                                                                                                                    \\ \hline
		Input  & States                                                                                                                                                                  & Latent variables                                                                                                                                                                           \\ \hline
		Arch.  & \begin{tabular}[c]{@{}l@{}}3$\times$3 Conv. 32, ReLU\\ 3$\times$3 Conv. 32, ReLU\\ 3$\times$3 Conv. 32, ReLU\\ 3$\times$3 Conv. 32\\ Flatten\\ Dense 512 \& Dense 512\\ Gaussian sampling\end{tabular} & \begin{tabular}[c]{@{}l@{}}Dense 64, ReLU\\ Dense 1024, ReLU\\ Reshape\\ 3$\times$3 Deconv. 64, ReLU\\ 3$\times$3 Deconv. 64, ReLU\\ 3$\times$3 Deconv. 64, ReLU\\ 8$\times$8 Deconv. 32\\ 1$\times$1 Conv. 4\end{tabular} \\ \hline
		Output & Latent variables                                                                                                                                                                                                & Predicted states                                                                                                                                                                \\ \hline
	\end{tabular}
\end{table}

To handle the graphic observations, we leveraged convolutional neural networks (CNNs) to build RISE and the benchmarking methods. For fair comparison, the same policy network and value network are employed for all the algorithms, and their architectures can be found in Table~\ref{tb:cnn na}. For instance, ``8$\times$8 Conv. 32" represents a convolutional layer that has $32$ filters of size 8$\times$8. A categorical distribution was used to sample an action based on the action probability of the stochastic policy. The VAE block of RISE and Max\Ry need to learn an encoder and a decoder. The encoder is composed of four convolutional layers and one dense layer, in which each convolutional layer is followed by a batch normalization (BN) layer \cite{ioffe2015batch}. Note that ``Dense 512 \& Dense 512" in Table~\ref{tb:cnn na} means that there are two branches to output the mean and variance of the latent variables, respectively. For the decoder, it utilizes four deconvolutional layers to perform upsampling while a dense layer and a convolutional layer are employed at the top and the bottom of the decoder. Finally, no BN layer is included in the decoder and the ReLU activation function is employed for all components.

\begin{figure*}[ht]
	\centering
	\includegraphics[width=0.95\linewidth]{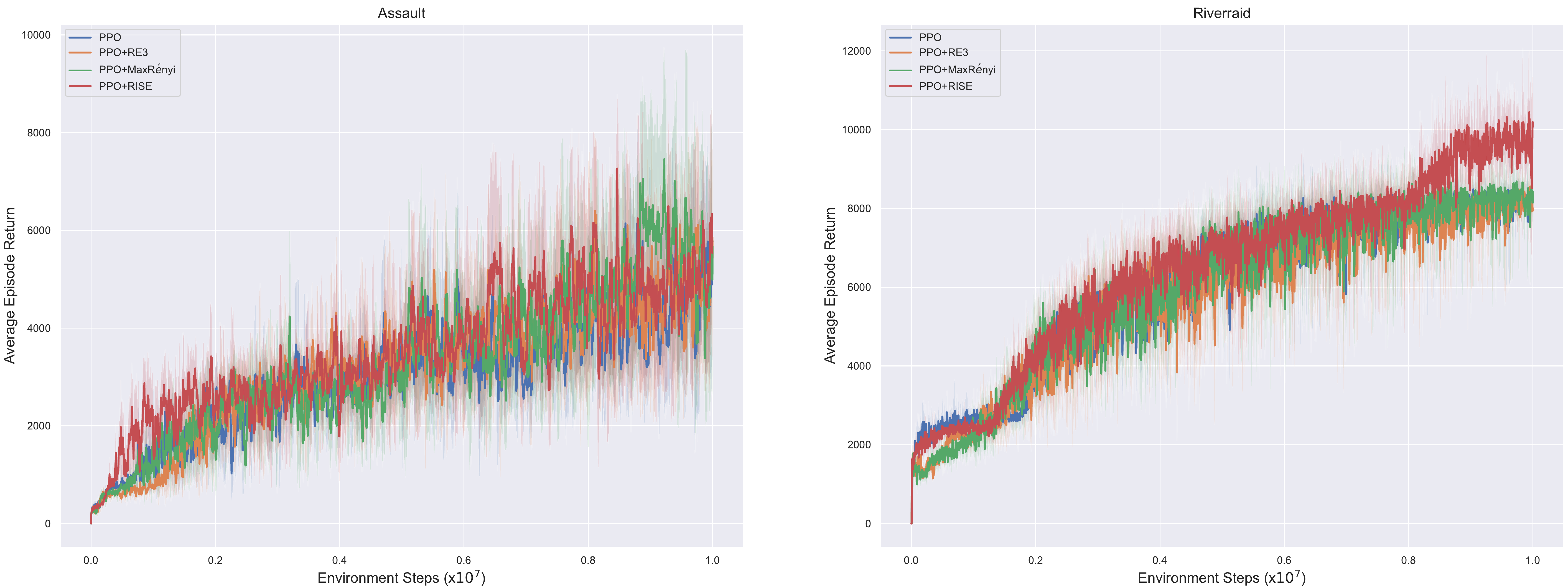}
	\caption{Average episode return versus number of environment steps on Atari games.}
	\label{fig:dis eps return}
\end{figure*}

\begin{table*}[h]
	\centering
	\caption{Performance comparison in nine Atari games.}
	\label{tb:dis final performance}
	\normalsize
	\begin{tabular}{l|l|l|l|l}
		\hline
		Game     & PPO     & PPO+RE3 & PPO+Max\Ry & PPO+RISE \\ \hline
		Assault  & 5.24k$\pm$1.86k   & 5.54k$\pm$2.12k   & 5.68k$\pm$1.99k   & \textbf{5.78k$\pm$2.44k}  \\
		Battle Zone & 18.63k$\pm$1.96k   & 20.47k$\pm$2.73k   & 19.17k$\pm$3.48k   & \textbf{21.80k$\pm$5.44k}  \\
		Demon Attack  & 13.55k$\pm$4.65k   & 17.40k$\pm$9.63k   & 16.18k$\pm$8.19k   & \textbf{18.39k$\pm$7.61k} \\
		Kung Fu Master & 21.86k$\pm$8.92k   & 23.21k$\pm$5.74k   & 27.20k$\pm$5.59k   & \textbf{27.76k$\pm$5.86k} \\
		Riverraid    & 7.99k$\pm$0.53k   & 8.14k$\pm$0.37k   & 8.21k$\pm$0.36k   & \textbf{10.07k$\pm$0.77k} \\
		Space Invaders & 0.72k$\pm$0.25k   & 0.89k$\pm$0.33k   & 0.81k$\pm$0.37k   & \textbf{1.09k$\pm$0.21k}\\
		\hline
	\end{tabular}
\end{table*}

In the first phase, we initialized a policy network $\pi_{\bm \theta}$ and let it interact with eight parallel environments with different random seeds. We first collected observation data over ten thousand environment steps before the VAE encoder generates the latent vectors of dimension of $128$ from the observation data. After that, the latent vectors were sent to the decoder to reconstruct the observation tensors. For parameters update, we used an Adam optimizer with a learning rate of $0.005$ and a batch size of $256$. Finally, we divided the observation dataset into $K=8$ subsets before searching for the optimal $k$-value over the range of $[1,15]$ using Algorithm~\ref{algo:k search}.

Equipped with the learned $k$ and encoder $q_{\bm \phi}$, we trained RISE with ten million environment steps. In each episode, the agent was also set to interact with eight parallel environments with different random seeds. Each episode has a length of $128$ steps, producing $1024$ pieces of transitions. After that, we calculated the intrinsic reward for all transitions using Eq.~\eqref{eq:intrinsic reward}, where $\alpha=0.1, \lambda_{0}=0.1$. Finally, the policy network was updated using a proximal policy optimization (PPO) method \cite{schulman2017proximal}. More specifically, we used a PyTorch implementation of the PPO method, which can be found in \cite{kostrikov2018github}. The PPO method was trained with a learning rate of $0.0025$, a value function coefficient of $0.5$, an action entropy coefficient of $0.01$, and a generalized-advantage-estimation (GAE) parameter of $0.95$ \cite{schulman2015high}. In particular, a gradient clipping operation with threshold $[-5, 5]$ was performed to stabilize the learning procedure. As for benchmarking methods, we trained them following their default settings reported in the literature \cite{zhang2021exploration,seo2021state}.

\subsubsection{Performance Comparison}

The average one-life return is employed as the KPI in our performance comparison. Table~\ref{tb:dis final performance} illustrates the performance comparison over eight random seeds on nine Atari games. For instance, 5.24k$\pm$1.86k represents the mean return is $5.24$k and the standard deviation is $1.86$k. The highest performance is shown in bold. As shown in Table~\ref{tb:dis final performance}, RISE achieved the highest performance in all nine games. Both RE3 and Max\Ry achieved the second highest performance in three games. Furthermore, Fig.~\ref{fig:dis eps return} illustrates the change of average episode return during training of two selected games. It is clear that the growth rate of RISE is faster than all the benchmarking methods.

\begin{figure}
	\centering
	\includegraphics[width=\linewidth]{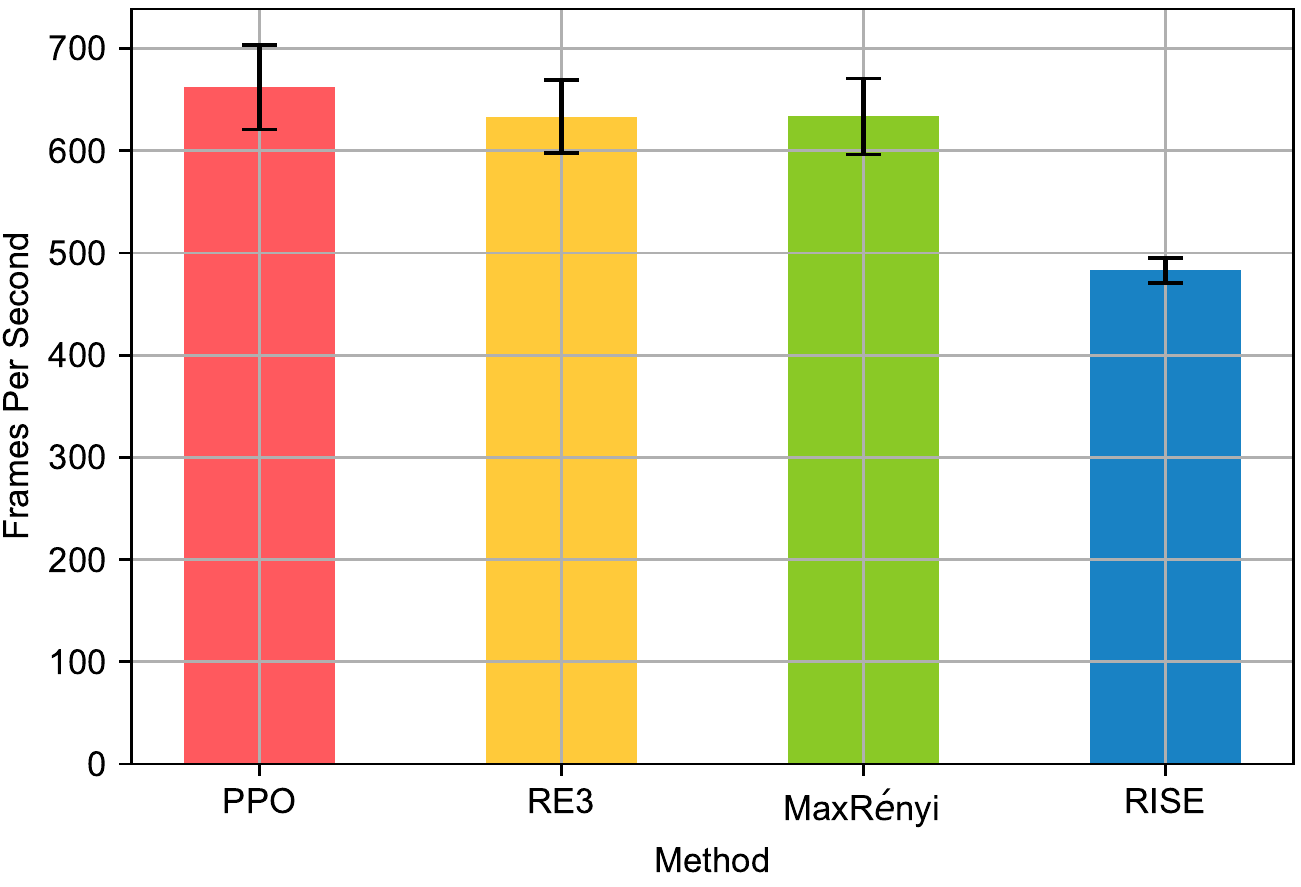}
	\caption{Average computational complexity on Atari games. The experiments were performed in Ubuntu 18.04 LTS operating system with a Intel 10900x CPU and a NVIDIA RTX3090 GPU.}
	\label{fig:fps}
\end{figure}

Next, we compare the training efficiency between RISE and the benchmarking methods, and the frame per second (FPS) is set as the KPI. For instance, if a method takes $10$ second to finish the training of an episode, the FPS is computed as the ratio between the time cost and episode length. The time cost involves only interaction and policy updates for the vanilla PPO agent. But the time cost needs to involve further the intrinsic reward generation and auxiliary model updates for other methods. As shown in Fig.~\ref{fig:fps}, the vanilla PPO method achieves the highest computation efficiency, while RISE and RE3 achieve the second highest FPS. In contrast, Max\Ry has far less FPS that RISE and RE3. This mainly because RISE and RE3 require no auxiliary models, while Max\Ry uses a VAE to estimate the probability density function. Therefore, RISE has great advantages in both the policy performance and learning efficiency.

\subsection{Bullet Games}
\subsubsection{Experimental Setting}
Finally, we tested RISE on six Bullet games \cite{coumans2016pybullet} with continuous action space, namely \textit{Ant}, \textit{Half Cheetah}, \textit{Hopper}, \textit{Humanoid}, \textit{Inverted Pendulum} and \textit{Walker 2D}. In all six games, the target of the agent is to move forward as fast as possible without falling to the ground. Unlike the Atari games that have graphic observations, the Bullet games use fixed-length vectors as observations. For instance, the ``Ant" game uses $28$ parameters to describe the state of the agent, and its action is a vector of $8$ values within $[-1.0, 1.0]$.

\begin{table*}[ht]
	\centering
	\caption{Performance comparison of six Bullet games.}
	\label{tb:con final performance}
	\normalsize
	\begin{tabular}{l|l|l|l|l}
		\hline
		Game     & PPO     & PPO+RE3 & PPO+Max\Ry & PPO+RISE \\ \hline
		Ant               & 2.25k$\pm$0.06k  & 2.36k$\pm$0.01k  & 2.43k$\pm$0.03k  & \textbf{2.71k$\pm$0.07k}  \\
		Half Cheetah      & 2.36k$\pm$0.02k &  2.41k$\pm$0.02k  & 2.40k$\pm$0.01k  & \textbf{2.47k$\pm$0.07k}  \\
		Hopper   		  & 1.53k$\pm$0.57k  & 2.08k$\pm$0.55k  & 2.23k$\pm$0.31k  &  \textbf{2.44k$\pm$0.04k}  \\
		Humanoid          &  0.83k$\pm$0.25k & 0.95k$\pm$0.64k  & 1.17k$\pm$0.54k &  \textbf{1.24k$\pm$0.92k}  \\
		Inverted Pendulum & 1.00k$\pm$0.00k  & 1.00k$\pm$0.00k  & 1.00k$\pm$0.00k  & \textbf{1.00k$\pm$0.00k} \\
		Walker 2D         & 1.66k$\pm$0.37k  & 1.85k$\pm$0.71k  & 1.73k$\pm$0.21k  & \textbf{1.96k$\pm$0.34k} \\
		\hline
	\end{tabular}
\end{table*}

\begin{table}[h]
	\caption{The MLP-based network architectures.}
	\label{tb:mlp na}
	\centering
	\begin{tabular}{l|l|l}
		\hline
		\textbf{Module} & Policy network $\pi_{\bm \theta}$                                                                                                                                                                                                 & Value network                                                                                                                                                                    \\ \hline
		Input  & States                                                                                                                                                                                                & States                                                                                                                                                                           \\ \hline
		Arch.  & \begin{tabular}[c]{@{}l@{}}Dense 64, Tanh\\ Dense 64, Tanh\\ Dense $|\mathcal{A}|$\\ Gaussian Distribution\end{tabular} & \begin{tabular}[c]{@{}l@{}}Dense 64, Tanh\\ Dense 64, Tanh\\ Dense 1 \end{tabular}                     \\ \hline
		Output & Actions                                                                                                                                                                                               & Predicted values                                                                                                                                                                           \\ \hline
		\textbf{Module} & Encoder $p_{\bm \psi}$                                                                                                                                                                                                       & Decoder $q_{\bm \phi}$                                                                                                                                                    \\ \hline
		Input  & States                                                                                                                                                                  & Latent variables                                                                                                                                                                           \\ \hline
		Arch.  & \begin{tabular}[c]{@{}l@{}}Dense 32, Tanh\\ Dense 64, Tanh\\ Dense 256\\ Dense 256 \& Dense 512\\ Gaussian sampling\end{tabular} & \begin{tabular}[c]{@{}l@{}}Dense 32, Tanh\\ Dense 64, Tanh\\ Dense observation shape\end{tabular} \\ \hline
		Output & Latent variables                                                                                                                                                                                                & Predicted states                                                                                                                                                                \\ \hline
	\end{tabular}
\end{table}


We leveraged the multilayer perceptron (MLP) to implement RISE and the benchmarking methods. The detailed network architectures are illustrated in Table~\ref{tb:mlp na}. Note that the encoder and decoder were designed for MaxR\'enyi, and no BN layers were introduced in this experiment. Since the state space is far simpler than the Atari games, the entropy can be directly derived from the observations while the training procedure for the encoder is omitted. We trained RISE with ten million environment steps. The agent was also set to interact with eight parallel environments with different random seeds, and Gaussian distribution was used to sample actions. The rest of the updating procedure was consistent with the experiments of the Atari games.

\subsubsection{Performance Comparison}
Table~\ref{tb:con final performance} illustrates the performance comparison between RISE and the benchmarking methods. Inspection of Table~\ref{tb:con final performance} suggests that RISE achieved the best performance in all six games. In summary, RISE has shown great potential for achieving excellent performance in both discrete and continuous control tasks.

\section{Conclusion}
In this paper, we have investigated the problem of improving exploration in RL by proposing a \Ry state entropy maximization method to provide high-quality intrinsic rewards. Our method generalizes the existing state entropy maximization method to achieve higher generalization capability and flexibility. Moreover, a $k$-value search algorithm has been developed to obtain efficient and robust entropy estimation by leveraging a VAE model, which makes the proposed method practical for real-life applications. Finally, extensive simulation has been performed on both discrete and continuous tasks from the Open AI Gym library and Bullet library. Our simulation results have confirmed that the proposed algorithm can substantially outperform conventional methods through efficient exploration.

\appendix
\subsection{Benchmarking Methods}\label{appendix:benchmarks}

\subsubsection{RE3}
Given a trajectory $\tau=(\bm{s}_{0},\bm{a}_{0},\dots,\bm{a}_{T-1},\bm{s}_{T})$, RE3 first uses a randomly initialized DNN to encode the visited states. Denote by $\{\bm{x}_{i}\}_{i=0}^{T-1}$ the encoding vectors of observations, RE3 estimates the entropy of state distribution $d(\bm{s})$ using a $k$-nearest neighbor entropy estimator \cite{singh2003nearest}:
\begin{equation}\label{eq:knn ee}
\begin{aligned}
\hat{H}_{T}^{k}(d) &= \frac{1}{T}\sum_{i=0}^{T-1}\log \frac{T\cdot \Vert \bm{x}_{i} - \tilde{\bm{x}}_{i} \Vert_{2}^{m} \cdot \pi}{k\cdot\Gamma(\frac{m}{2}+1)}+\log k-\Psi(k) \\
&\propto \frac{1}{T}\sum_{i=0}^{T-1}\log\Vert \bm{x}_{i} - \tilde{\bm{x}}_{i} \Vert_{2},
\end{aligned}
\end{equation}
where $\tilde{\bm{x}}_{i}$ is the $k$-nearest neighbor of $\bm{x}_{i}$ within the set $\{\bm{x}_{i}\}_{i=0}^{T-1}$, $m$ is the dimension of the encoding vectors, and $\Gamma(\cdot)$ is the Gamma function, and $\Psi(\cdot)$ is the digamma function. Note that $\pi$ in Eq.~\eqref{eq:knn ee} denotes the ratio between the circumference of a circle to its diameter. Equipped with Eq.~\eqref{eq:knn ee}, the total reward for each transition $(\bm{s}_{t},\bm{a}_{t},r_{t},\bm{s}_{t+1})$ is computed as:
\begin{equation}
r^{\rm total}=r(\bm{s}_{t},\bm{a}_{t})+\lambda_{t} \cdot \log(\Vert \bm{x}_{t} - \tilde{\bm{x}}_{t} \Vert_{2} + 1),
\end{equation}
where $\lambda_{t}=\lambda_{0}(1-\kappa)^{t},\lambda_{t}\geq 0$ is a weight coefficient that decays over time, $\kappa$ is a decay rate. Our RISE method is a generalization of RE3, which provides more aggressive exploration incentives.

\subsubsection{Max\Ry}
The May\Ry method aims to maximize the \Ry entropy of state-action distribution $d^{\pi}_{\rho}(\bm{s},\bm{a})$. The gradient of its objective function is
\begin{equation}
	\begin{aligned}
		\nabla_{\bm{\theta}} H_{\alpha}(d^{\pi}_{\rho})\propto\frac{\alpha}{1-\alpha}\mathbb{E}_{(\bm{s},\bm{a})\sim d^{\pi}}\bigg[\nabla_{\bm{\theta}}\log\pi(\bm{a}|\bm{s})\\
		\left(\frac{1}{1-\gamma}\langle d_{\bm{s},\bm{a}}^{\pi}, (d^{\pi}_{\rho})^{\alpha-1} \rangle+(d^{\pi}_{\rho}(\bm{s},\bm{a}))^{\alpha-1}\right)\bigg].
	\end{aligned}
\end{equation}
It uses VAE to estimate $d(\bm{s})$ and take the evidence lower bound (ELBO) as the density estimation \cite{kingma2013auto}, which suffers from low efficiency and high variance.


\subsection{Proof of Lemma \ref{lemma:smooth}}\label{appendix:proof of lemma}
Since $\nabla^{2}\tilde{H}_{\alpha,\sigma}(d)=-\alpha(d(\bm{s})+\sigma)^{\alpha-2}$ is a diagonal matrix, we have
\begin{equation}
	\begin{aligned}
		&\Vert\nabla\tilde{H}_{\alpha,\sigma}(d)-\nabla\tilde{H}_{\alpha,\sigma}(d')\Vert_{\infty}\\
		&\leq\max_{\varsigma\in[0,1]}|\nabla^{2}\tilde{H}_{\alpha,\sigma}(\varsigma d+(1-\varsigma)d')|\cdot\Vert d-d'\Vert_{\infty}\\
		&\leq \alpha\sigma^{\alpha-2}\Vert d-d'\Vert_{\infty},
	\end{aligned}
\end{equation}
where the first inequality follows the Taylor's theorem. This concludes the proof.

\subsection{Proof of Theorem \ref{theorem:sample complexity}}\label{appendix:proof of theorem}
Equipped with Lemma \ref{lemma:smooth}, let $\pi^{*}=\underset{\pi\in\Pi}{\rm argmax}\:\tilde{H}_{\alpha,\sigma}(d)$, we have (proved by \cite{hazan2019provably}):
\begin{equation}
	\begin{aligned}
		\tilde{H}_{\alpha,\sigma}(d^{\pi^*})-\tilde{H}_{\alpha,\sigma}(d^{\pi_{{\rm mix},T}})\\
		\leq B\exp(-T\eta)+2\beta\epsilon_2+\epsilon_1+\eta\beta,
	\end{aligned}
\end{equation}
where $\Vert\nabla\tilde{H}_{\alpha,\sigma}(d)\Vert_{\infty}\leq B=\frac{\alpha}{1-\alpha}\sigma^{\alpha-1}$. Thus it suffices to set $\epsilon_1=0.1\epsilon,\epsilon_2=0.1\beta^{-1}\epsilon,\eta=0.1\beta^{-1}\epsilon,T=\eta^{-1}\log10B\epsilon^{-1}$. When Algorithm \ref{algo:mpec} is run for 
\begin{equation}
	T\geq 10\beta\epsilon^{-1}\log10B\epsilon^{-1},
\end{equation}
it holds
\begin{equation}
	\tilde{H}_{\alpha,\sigma}(d^{\pi_{{\rm mix},T}})\geq \max_{\pi\in\Pi}\tilde{H}_{\alpha,\sigma}(d^{\pi})-\epsilon.
\end{equation}
Consider the imposed smoothing on $\tilde{H}$, we set $\sigma$ and
This concludes the proof.

\subsection{Off-policy Version of RISE}
\begin{algorithm}[h]
	\caption{Off-policy RL Version of RISE}
	\label{algo:rise off-policy}
	\begin{algorithmic}[1]
		\STATE \textbf{Phase 1} of Algorithm \ref{algo:rise on-policy};
		\STATE \textbf{Phase 2: Policy update}
		\STATE Initialize the maximum environment steps $t_{\rm max}$, order $\alpha$, coefficients $\lambda_{0},\zeta$, decay rate $\kappa$, and replay buffer $\mathcal{B}\leftarrow \emptyset$;
		
		\FOR {step $t=1,\dots,t_{\rm max}$}
		\STATE Collect the transition $(\bm{s}_{t},\bm{a}_{t},r_{t},\bm{s}_{t+1})$ and let $\mathcal{B}\leftarrow \mathcal{B}\cup\{(\bm{s}_{t},\bm{a}_{t},r_{t},\bm{s}_{t+1},\bm{z}_{t})\}$, where $\bm{z}_{t}=q_{\bm \theta}(\bm{s}_{t})$;
		\STATE Sample a minibatch $\{(\bm{s}_{i},\bm{a}_{i},r_{i},\bm{s}_{i+1},\bm{z}_{i})\}_{i=1}^{B}$ from $\mathcal{B}$ randomly;
		\STATE Compute $\hat{r}(\bm{s}_{i})\leftarrow (\Vert \bm{z}_{i}-\tilde{\bm{z}}_{i} \Vert_{2})^{1-\alpha}$;
		\STATE Update $\lambda_{t}=\lambda_{0}(1-\kappa)^{t}$;
		\STATE Let $r^{\rm total}_{i}=r(\bm{s}_{i},\bm{a}_{i})+\lambda_{t} \cdot \hat{r}(\bm{s}_{i})+\zeta\cdot H(\pi(\cdot|\bm{s}_i))$;
		\STATE Update the policy network with transitions $\{\bm{s}_{i},\bm{a}_{i},\bm{s}_{i+1},r^{\rm total}_{i}\}_{i=1}^{B}$ using any off-policy RL algorithms.
		\ENDFOR
	\end{algorithmic}
\end{algorithm}

\end{document}